\pdfoutput=1

\documentclass[11pt]{article}

\usepackage{acl}

\usepackage{times}
\usepackage{latexsym}
\usepackage{graphicx}
\usepackage{amsmath}
\usepackage{booktabs}
\usepackage{multirow}
\usepackage{url}
\usepackage{enumitem}
\usepackage{microtype}
\usepackage{subcaption}
\usepackage{makecell}
\usepackage{lipsum}
\usepackage{hyperref}
\usepackage{verbatim}
\usepackage{natbib}
\usepackage{pifont}

\definecolor{darkgreen}{RGB}{0,100,0}
\definecolor{GREEN}{RGB}{84,130,53}
\newcommand{\chkmark}{\ding{51}}%
\newcommand{\crsmark}{\ding{55}}%

\newcommand{\system}{\textsc{ShortCheck}}

\title{\system: Checkworthiness Detection of Multilingual Short-Form Videos} 

\author{
Henrik Vatndal \\
Factiverse AI\\
\texttt{henrik@factiverse.ai}
\And
Vinay Setty \\
Factiverse AI and University of Stavanger \\
\texttt{vsetty@acm.org}
}

\begin{document}
\maketitle

\begin{abstract}
Short-form video platforms like TikTok present unique challenges for misinformation detection due to their multimodal, dynamic, and noisy content. 
We present \system, a modular, inference-only pipeline with a user-friendly interface that automatically identifies checkworthy short-form videos to help human fact-checkers.
The system integrates speech transcription, OCR, object and deepfake detection, video-to-text summarization, and claim verification. 
\system~is validated by evaluating it on two manually annotated datasets with TikTok videos in a multilingual setting.
The pipeline achieves promising results with F1-weighted score over 70\%. 
\end{abstract}

\section{Introduction}

\begin{figure}[t!!]
\centering
\begin{minipage}[t]{0.5\textwidth}
    \centering
    \includegraphics[width=0.5\linewidth]{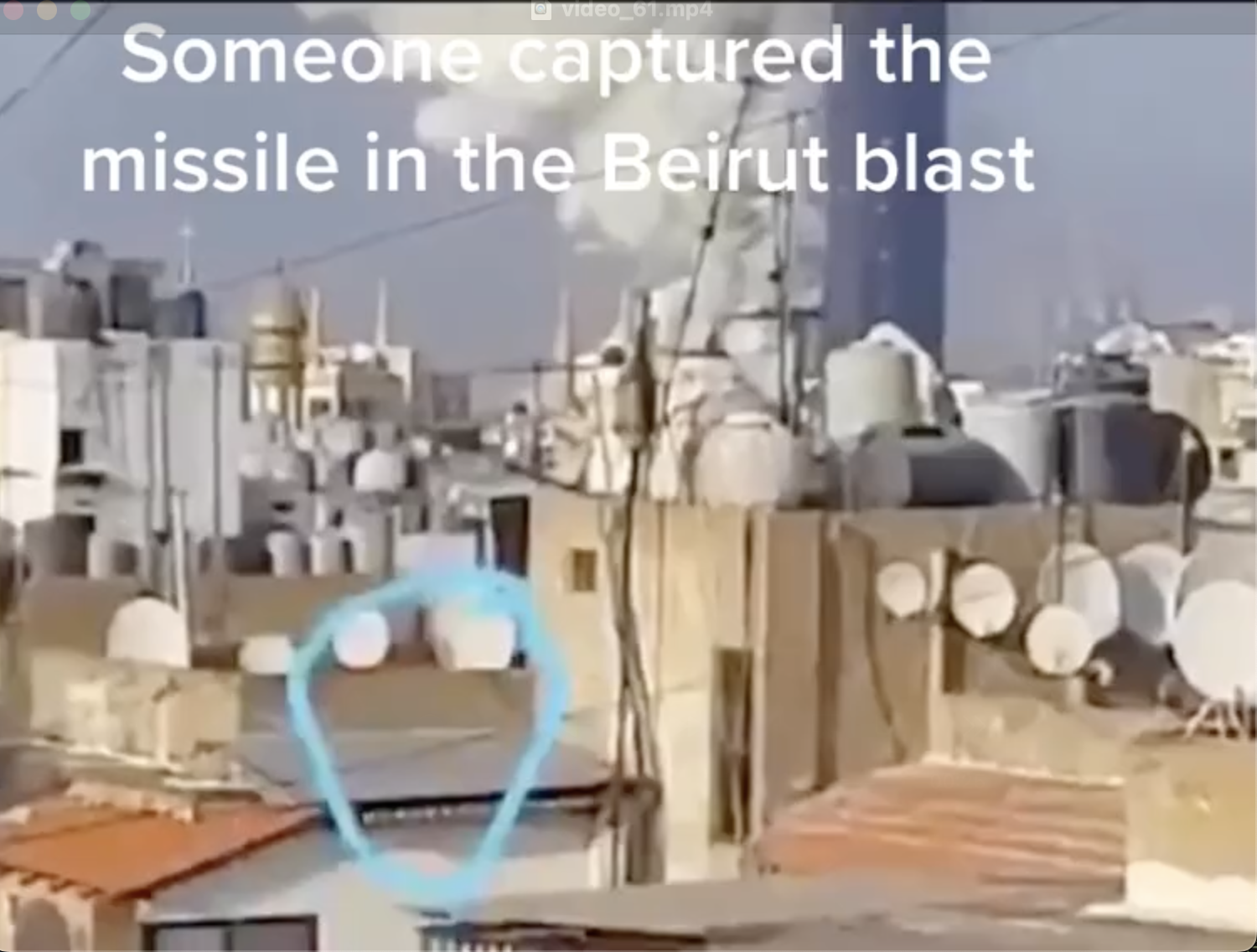}
    
\end{minipage}\hfill
\begin{minipage}[t]{0.5\textwidth}
    \centering
    \small
    \begin{tabular}{lp{4cm}}
    \midrule
    \textbf{Field} & \textbf{Value} \\ 
    \midrule
    overlay\_text & Someone captured the \textbar{} missile in the Beirut blast \\ \hline
    transcript & It's a shame (in Arabic)\\
    \hline
    video\_summary & The video captures footage of the 2020 Beirut blast, showing destruction and chaos in an urban area, with explosions visible throughout. \\ \hline
    buzzword\_detected & False \\ \hline
    transcript\_verdict & hostile \\ \hline
    summary\_verdict & contentious-issue \\\hline
    overlay\_verdict & hostile \\ \hline
    Checkworthiness & \textcolor{darkgreen}{True} \\\hline
    \midrule
    \end{tabular}
\end{minipage}
\caption{\small TikTok video with overlay text claiming ``Someone captured the missile in the Beirut blast,'' and noisy Arabic audio and explosion visuals. Features below show why \system~marked it \emph{Checkworthy}.}
\label{fig:example}
\end{figure}

The popularity of short-form video platforms such as \emph{TikTok}, \emph{YouTube Shorts} and \emph{Instagram Reels} has transformed how information is produced, consumed, and spread. 
With billions of monthly active users, these platforms create fertile ground for the spread of misinformation on sensitive topics including politics, health, and social issues. 
Unlike traditional text or image-based content, these videos may include multiple modalities such as speech, text overlays, music, and visuals, often edited in ways that obscure meaning or context, though not all of these elements are always present; for example, some videos contain only on-screen text without audio, while others show just a speaker without any additional graphics or overlays.
For example, Figure \ref{fig:example} shows a TikTok screenshot where the overlay text makes the claim, while the audio transcript (translated to ``it’s a shame'') does not provide any useful information. The video summary notes an urban explosion, indicating potentially contentious content for fact-checkers.

The multimodal complexity of short videos makes automated fact-checking technically challenging and manual efforts are increasingly unsustainable, especially for under-resourced fact-checkers facing unprecedented content scale and funding cuts.
In this demo, we present a prototype designed to automate the identification of potentially checkworthy videos, significantly reducing the time required by human fact-checkers.
Our prototype is easy to use and can predict checkworthiness in over 30 major languages.\footnote{Intersection of languages supported by Meta Llama3 \url{https://ai.meta.com/blog/meta-llama-3/} and OpenAI Whisper \url{https://github.com/openai/whisper}}

\begin{table*}[ht!!]
\centering
\small
\caption{Fact-checking systems categorized by input modality, granularity, multilinguality, and  fact-checking support.}
\begin{tabular}{lccccccc}
\toprule
\textbf{System} & \textbf{Text} & \textbf{Audio} &  \textbf{Image} &\textbf{Video} & \textbf{Granularity} & \textbf{M.lingual} & \textbf{Full FC} \\
\midrule
\textsc{BRENDA}~\citet{Botnevik:2020:SIGIR} & \chkmark & \crsmark & \crsmark & \crsmark & Long Text & \crsmark & \chkmark \\
\textsc{FLEEK}~\cite{Bayat:2023:EMNLP} & \chkmark & \crsmark & \crsmark & \crsmark & Single Claim & \crsmark & \chkmark \\
\textsc{QACheck}~\cite{Pan:2023:EMNLP} & \chkmark & \crsmark & \crsmark& \crsmark & Single Claim & \crsmark & \chkmark \\
\textsc{ClaimLens}~\cite{Devasier:2024:EMNLP} & \chkmark & \crsmark & \crsmark& \crsmark & Single Claim & \crsmark & \crsmark \\
\textsc{TruthReader}~\cite{Ming:2024:EMNLP} & \chkmark & \crsmark & \crsmark& \crsmark & Long Text & \crsmark & \chkmark \\
\textsc{OpenFactCheck}~\cite{Zhou:2024:EMNLP} & \chkmark & \crsmark & \crsmark& \crsmark & Long Text & \crsmark & \chkmark \\
\textsc{FactCheckEditor}~\cite{Setty:2024:SIGIR} & \chkmark & \crsmark& \crsmark & \crsmark & Long Text & \chkmark & \chkmark \\
\textsc{Loki}~\cite{Li:2025:COLING} & \chkmark & \crsmark& \crsmark & \crsmark & Single Claim & \chkmark & \chkmark \\
\textsc{AudioCWD}~\cite{Ivanov:2024:ICASSP} & \crsmark & \chkmark & \crsmark& \crsmark & Single Claim & \crsmark & \crsmark \\
\textsc{LiveFC}~\cite{Venktesh:2025:WSDM} & \chkmark & \chkmark & \crsmark& \crsmark & Long Text & \crsmark & \chkmark \\
\textsc{PodFC}~\cite{Setty:2025:WSDM} & \chkmark & \crsmark& \chkmark & \crsmark & Long Text & \chkmark & \chkmark \\
\textsc{Fauxtography}~\cite{Zlatkova:2019:EMNLP} & \chkmark & \crsmark & \chkmark& \crsmark & Single Claim & \crsmark & \chkmark \\
\textsc{AVerImaTeC}~\cite{Cao:2025:arXiv} & \chkmark & \crsmark & \chkmark&  \crsmark & Single Claim & \crsmark & \chkmark \\
\textsc{CER}~\cite{Zuo:2025:SIGIR} & \chkmark & \chkmark & \chkmark & \chkmark & Single Claim & \crsmark & \chkmark \\
\textsc{COVID-VTS}~\cite{Liu:2023:EACL} & \chkmark & \chkmark & \chkmark & \chkmark & Single Claim & \crsmark & \crsmark \\
\system~(ours) & \chkmark & \chkmark & \chkmark &  \chkmark & Long Text & \chkmark & \crsmark \\
\bottomrule
\end{tabular}

\label{tab:modality_factcheck}
\vspace{-10pt}
\end{table*}

Most existing misinformation detection systems are designed for structured, single-modality content such as news articles, social media posts, or deepfake detection, with a primary focus on either text, audio, transcriptions or visual modalities. We summarize the existing fact-checking systems and their modalities in Table \ref{tab:modality_factcheck}. However, these approaches are not suited for short-form video content found on platforms like TikTok or YouTube Shorts due to the casual unstructured nature of the content.

Short-form videos pose unique challenges due to their \textit{limited multimodal generalization}. They often blend speech, text, music, and visuals in non-linear, asynchronous ways that traditional unimodal models struggle to interpret \cite{Alam:2022:COLING, Yao:2023:SIGIR, Guo:2022:TACL, Singhal:2019:BigMM}. These videos also exhibit \textit{noisy or incomplete modality signals}: some lack audio, others feature distorted overlays or rapid cuts, making existing detectors brittle in real-world conditions \cite{Jindal:2020:SafeAI, Venktesh:2024:SIGIR}. Furthermore, most models offer \textit{low interpretability}, returning binary predictions without justifications. This hinders adoption in professional workflows that require transparent, evidence-backed reasoning \cite{Schlichtkrull:2023:NeurIPS, Guo:2022:TACL, Alam:2022:COLING, Venktesh:2025:WSDM}.

While multimodal fact-checking is gaining traction, the gap between research prototypes and deployable, interpretable tools for short-form video remains substantial. Bridging this divide requires not only improved multimodal understanding but also system outputs that align with the needs of human fact-checkers in high-throughput environments.

\paragraph{Our Contributions.}
This paper introduces a demonstration system for detecting checkworthy TikTok videos. 
Our key contributions are:
\begin{itemize}[noitemsep]
    \item A \textbf{modular multimodal pipeline} that integrates OCR, transcription, video-to-text captioning, semantic classification, retrieval and fact-checking modules.
    \item Two \textbf{new multilingual annotated datasets} of TikTok videos labeled for checkworthiness.
    \item An \textbf{evaluation and error analysis} showing which modalities contribute most to reliable classification.
    \item A \textbf{demo interface} that allows fact-checkers to upload videos, inspect intermediate results, and link claims to existing fact-checks. 
\end{itemize}
Together, these contributions provide a step toward bridging the gap between state-of-the-art research and practical tools for combating misinformation on emerging short-form video platforms.

\section{Related work}
\label{sec:related}

\paragraph{Text-Based Fact-Checking and Datasets.}
Automated fact-checking has progressed with the rise of benchmark datasets. The FEVER dataset by \citet{Thorne:2018:NAACL} established foundational tasks in claim verification, with follow-up surveys and taxonomies by \citet{Thorne:2018:COLING} and \citet{Guo:2022:TACL}. Real-world datasets such as \textsc{MultiFC} \citep{Augenstein:2019:EMNLP}, \textsc{AVeriTeC} \citep{Schlichtkrull:2023:NeurIPS}, and \textsc{QuanTemp} \citep{Venktesh:2024:SIGIR} have expanded the scope to include diverse, evidence-backed, and numerically grounded claims.

\paragraph{CheckWorthiness.}
To scale verification, identifying check-worthy claims is critical. \citet{Gencheva:2017:RANLP} introduced context-aware ranking, while \textsc{ClaimRank} by \citet{Jaradat:2018:NAACL} provides real-time claim ranking for journalists.

\paragraph{Multimodal and Video-Centric Fact-Checking.}
Multimodal misinformation has led to techniques that fuse text, audio, and visual cues \citep{Alam:2022:COLING}. \textsc{FakingRecipe} \citep{Bu:2024:ICMM} investigates manipulation strategies in TikTok videos, complementing earlier datasets like \textsc{SpotFake} \citep{Singhal:2019:BigMM} and \textsc{NewsBag} \citep{Jindal:2020:SafeAI}. \citet{Yao:2023:SIGIR} proposed an end-to-end video fact-checking system with explanation generation. Our work builds on these by offering a modular, interpretable pipeline designed for integration into professional fact-checkers’ workflows.

\paragraph{Deepfake Detection.}
Deepfake detection models like \textsc{MesoNet} \citep{Afchar:2018:WIFS} target facial forgeries using lightweight CNNs, making them suitable for integration into broader video authenticity pipelines.

\paragraph{Practical Systems and Live Fact-Checking.}
Recent systems have bridged research and application by supporting live or end-to-end fact-checking across modalities. \textsc{BRENDA} \citep{Botnevik:2020:SIGIR} provides real-time claim detection for long-form text. \textsc{FactCheckEditor} \citep{Setty:2024:SIGIRa} and \textsc{PodFC} \citep{Setty:2025:WSDM} support multilingual verification, with the latter tailored for long-form audio. \textsc{LiveFC} \citep{Venktesh:2025:WSDM} handles real-time fact-checking of audio streams, combining transcription, claim detection, evidence retrieval, and verdict generation. However, as summarized in Table~\ref{tab:modality_factcheck}, most systems focus on single-modality inputs (usually text) and single-claim granularity, with limited support for video or multimodal content.

Our system, \system, differs in its emphasis on short-form video platforms like TikTok. It uniquely supports all four modalities: text, audio, image, and video, along with multilingual processing and capable of processing long text with more than single claims. While it does not support full automated fact-checking, it offers interpretable, modular verdict signals suitable for assisting professional fact-checkers in verification.

\section{System Overview}
\label{sec:method}

\begin{figure*}[h!]
    \centering
    \includegraphics[width=\textwidth]{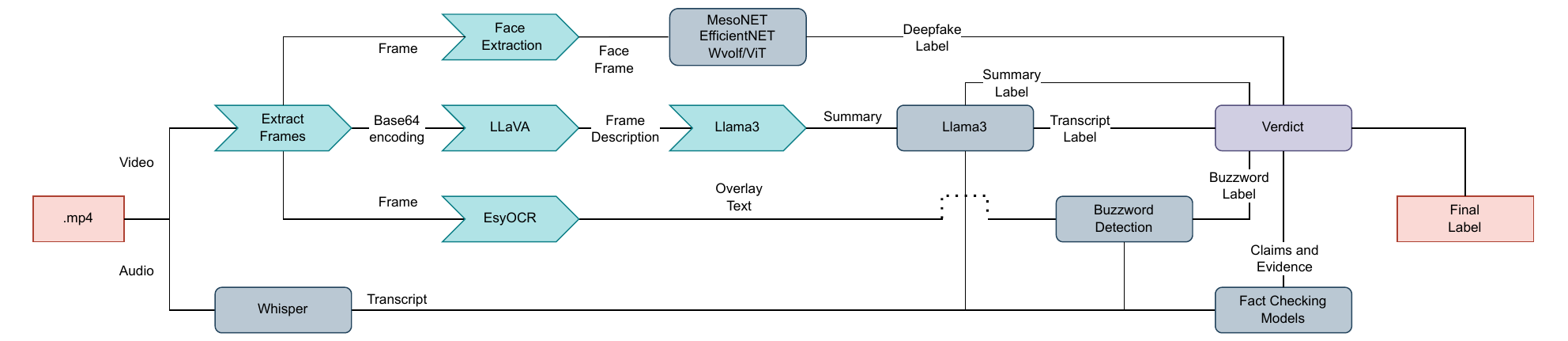}
    \caption{Modular pipeline for fact-checking TikTok videos.}
    \label{fig:pipeline}
\end{figure*}

Given that the checkworthiness of a TikTok video can be inherently subjective, we base our approach on established best practices followed by professional fact-checkers. In particular, we consulted the guidelines of Faktisk.no\footnote{A Norwegian non-profit organization accredited by the International Fact-Checking Network (IFCN).} This also aligns with the definition of fact-checkworthiness in the literature\cite{Jaradat:2018:NAACL, Barron:2024:CLEF}

\begin{enumerate}
    \item A short video is defined as up to 10 minutes long.
    \item A short video can contain one or more claims, often complex interaction between modalities.
    \item Checkworthiness is defined as videos which have potential to cause harm to the public particularly related to politics, health and society. 
    \item Claims related to celebrity gossip, sports, advertisements etc are not considered checkworthy.
\end{enumerate}

We propose a modular, inference-only pipeline for detecting potentially misleading or checkworthy content in short-form videos, particularly those published on platforms like TikTok. The system assigns each video one of two categorical labels: \texttt{Checkworthy}, or \texttt{Not\_Checkworthy}. Unlike prior work that builds monolithic or end-to-end models, our design emphasizes modularity, interpretability, and adaptability. Each component in the pipeline can be independently replaced, which makes the system robust to failures in specific modalities and easier to maintain in production settings.

\subsection{Pipeline Components}

The pipeline comprises feature extraction modules tailored to speech, text, visuals, and metadata, whose outputs are aggregated by a rule-based engine for final video classification. Additional modules, such as object detection for weapons, were tested but excluded due to limited contribution.

\paragraph{Optical Character Recognition (OCR):}
The first stage involves extracting visible on-screen text through Optical Character Recognition (OCR), using the EasyOCR library\footnote{\url{https://github.com/JaidedAI/EasyOCR}}. This module captures embedded captions or textual overlays, which are common in TikTok videos. However, OCR performance is often challenged by stylized fonts, rapid transitions, and visually noisy frames.

\paragraph{DeepFake Detection:}
To detect synthetic media, we incorporated a deepfake detection module into the pipeline. Since many methods rely on identity-specific data or high-quality frontal imagery, we evaluated their generalization to TikTok’s unconstrained, user-generated content \cite{abbas2024unmasking_deepfake_survey}. We tested three zero-shot models: \textit{MesoNet} \cite{afchar2018mesonet}, a mesoscopic CNN; \textit{EfficientNet} \cite{9412711_icpr_efficient,dolhansky2020dfdc}, an attention-based model from the DFDC challenge; and \textit{Wvolf/ViT}\footnote{\url{https://huggingface.co/Wvolf/ViT_Deepfake_Detection}} \cite{bonettini2020videofacemanipulationdetection}, selected for its plug-and-play accessibility. Their outputs contributed as signals in the rule-based decision logic.

\paragraph{Speech Transcription:}
Speech transcription is handled by OpenAI Whisper \cite{Radford:2022:Whisper}, which offers robust multilingual transcription and performs well in noisy audio environments. However, overlapping music or sound effects, which are prevalent in entertainment-oriented videos, can still degrade transcription quality.

\paragraph{Video Summarization:}
To obtain a visual semantic summary, video frames are sampled and passed through LLaVA \cite{Liu:2023:LLaVA}, a vision-language model that generates frame-level captions describing people, objects, and scenes. These captions are subsequently summarized and contextualized using Meta’s LLaMA 3 model, which also performs high-level semantic classification. The model predicts whether a video is political, hostile, benign, or promotional in nature. All models are hosted via Ollama, a lightweight, local model serving platform that supports REST-based inference.\footnote{\url{ollama.com}}

\paragraph{Ideological Language Detection:}
In addition to these core modules, we incorporate a rule-based system for detecting ideological buzzwords and coded language, often referred to as dog whistles. This module operates on both OCR and transcript outputs, scanning for terms known to encode political or ideological meaning in subtle ways. The detection rules are informed by prior literature on dog-whistle communication \cite{Albertson:2015:DogWhistles} and curated datasets from organizations such as Faktisk.no.

\paragraph{Text-based fact-checking}
To further refine the decision-making process, we incorporate a claim detection and external fact-checking module. This component leverages fine-tuned transformer models for claim detection and natural language inference (NLI), applied to both the transcript and the visual summary of the video. Following the approach proposed by \cite{Setty:2024:SIGIRa}, the module identifies declarative, factual statements and attempts to verify them against a fact-checking evidence database. While this does not fact-check the entire content of the video, it provides fact-checkers early signals indicating whether the video may contain verifiably false or misleading information.

Finally, the system aggregates the outputs from all modules using a rule-based logic engine. A scoring system considers multiple module outputs, including the presence of claims,  ideological language, or political classification. If the cumulative score exceeds a threshold, the video is marked as \texttt{Checkworthy}. Advertisements are detected and filtered early in the pipeline and override all other scores. Videos that do not meet either criterion are labeled as \texttt{Not\_Checkworthy}. This decision process is entirely transparent and configurable, enabling future refinement without retraining models.

\subsection{Model Configuration and Deployment}

All models used in the pipeline are inference-only and require no task-specific fine-tuning. We use prompt engineering to adapt general-purpose models to specific sub-tasks. The LLaMA 3 and LLaVA models are deployed using Ollama, which allows for lightweight, local hosting and fast prototyping. Custom prompts, temperature settings, stop sequences, and token limits are adjusted to ensure consistent outputs across modules.

\subsection{Interpretability and Modularity}

A key design goal of our approach is interpretability. Each module exposes intermediate outputs that are human-readable and can be inspected by fact-checkers. This transparency builds trust and enables feedback-driven improvement of the system. The modular design also ensures that any individual component, such as the OCR engine or the semantic classifier, can be replaced or updated without disrupting the entire pipeline. This is especially important for deployment in evolving information ecosystems where content formats and threat types change rapidly.

\section{Experimental Evaluation}
\label{sec:expsetup}
\subsection{Setup}
All experiments were conducted on a local machine with an \textbf{NVIDIA A10 GPU (24GB VRAM)}. All models, including vision-language models and LLMs, were served locally via \texttt{Ollama} using REST-based endpoints.

\subsection{Datasets}

We evaluate \system~on two manually annotated dataset. Summary of dataset statistics is shown in Table~\ref{tab:dataset_summary}

\paragraph{Norwegian influencer data:} This dataset includes 249 TikTok videos curated by Faktisk for an emotional analysis study on political trolling via buzzwords like ``Stem FRP''.\footnote{\url{https://www.faktisk.no/artikkel/faktisk-analyse-av-tiktok-menn-mest-negative/109375}} We manually annotated each video for checkworthiness using the guidelines in Section~\ref{sec:method}.

\paragraph{TikTok Videos from Fact-Checking Websites:} This dataset, curated by \cite{Bu:2024:ICMM}, was compiled from fact-checking platforms including Snopes, PolitiFact, FactCheck.org, and Health Feedback. While the majority of content is in English, some posts and modalities appear in other languages. We annotated a sample of 254 videos from this collection, following the same guidelines described before.
\begin{table*}[ht!]
\centering
\small
\caption{Results on TikTok videos in Norwegian and English. CW = Checkworthy, NCW = Not Checkworthy. Combined metrics are macro-averages. Metrics: Precision (P), Recall (R), Accuracy (Acc) and F1-weighted (F1-W)}
\begin{tabular}{lcccccccccc}
\toprule
\textbf{Dataset} & \multicolumn{3}{c}{\textbf{CW}} & \multicolumn{3}{c}{\textbf{NCW}} & \multicolumn{4}{c}{\textbf{Combined (Macro)}} \\
\cmidrule(lr){2-4} \cmidrule(lr){5-7} \cmidrule(lr){8-11}
& P & R & F1-W & P & R & F1-W & P & R & F1-W & Acc \\
\midrule
Norwegian influencer & 0.64 & 0.85 & 0.73 & 0.95 & 0.92 & 0.93 & 0.74 & 0.73 & 0.72 & 0.88 \\
Fact-checking websites   & 0.82 & 0.58 & 0.68 & 0.72 & 0.90 & 0.80 & 0.77 & 0.74 & 0.74 & 0.76 \\
\bottomrule
\vspace{-10pt}
\end{tabular}

\end{table*}
\begin{table}[ht!]
\centering
\caption{Dataset composition by language and checkworthiness class. CW: Checkworthy and NCW: Not Checkworthy.}
\small
\begin{tabular}{lccc}
\toprule
\textbf{Dataset} & \textbf{CW} & \textbf{NCW} & \textbf{Total} \\
\midrule
Norwegian influencer & 33 & 204 & 237 \\
Fact-checking websites  & 114 & 140 & 254 \\
\midrule
\textbf{Total} & 147 & 344 & 491 \\
\bottomrule
\end{tabular}
\label{tab:dataset_summary}
\end{table}

\begin{table}[ht!]
\centering
\small
\caption{Evaluation scores of DeepFake detection models. Precision (P), Recall (R), Accuracy (A), and F1-weighted (F1-w)}
\begin{tabular}{lcccc}
\hline
\textbf{Model} & \textbf{A} & \textbf{P} & \textbf{R} & \textbf{F1-W} \\ \hline
MesoNET &      0.114 &       0.808 &    0.019 &      0.038 \\
Wvolf/ViT &      0.100 &       0.000 &    0.000 &      0.000 \\ 
EfficientNET &      \textbf{0.612} &       \textbf{0.992} &    \textbf{0.573} &      \textbf{0.727} \\
\hline
\end{tabular}
\end{table}

\begin{table}[ht!]
\centering
\small
\caption{Ablation study showing the performance change when individual modules are removed. Values represent the change from the full system (Baseline). Red indicates a drop in performance. Metrics: Precision (P), Recall (R), Accuracy (Acc) and F1-weighted (F1-W).}
\label{tab:ablation_deltas}
\begin{tabular}{lcccc}
\hline
\textbf{Removed} & \textbf{P} & \textbf{R} & \textbf{Acc} & \textbf{F1-W} \\
\hline
Weapon detection       & \textcolor{darkgreen}{+0.005} & \textcolor{darkgreen}{+0.002} & \textcolor{darkgreen}{+0.004} & \textcolor{darkgreen}{+0.004} \\
Video summary  & \textcolor{darkgreen}{+0.001} & \textcolor{red}{-0.008} & \textcolor{black}{0.000} & \textcolor{red}{-0.002} \\
Transcript      & \textcolor{darkgreen}{+0.027} & \textcolor{red}{-0.076} & \textcolor{black}{0.000} & \textcolor{red}{-0.024} \\
Buzzword               & \textcolor{red}{-0.024} & \textcolor{red}{-0.050} & \textcolor{red}{-0.021} & \textcolor{red}{-0.033} \\
OCR             & \textcolor{red}{-0.005} & \textcolor{red}{-0.030} & \textcolor{red}{-0.003} & \textcolor{red}{-0.004} \\
Fact Check             & \textcolor{darkgreen}{+0.013} & \textcolor{red}{-0.040} & \textcolor{darkgreen}{+0.004} & \textcolor{red}{-0.009} \\
\hline
All modules & \textbf{0.737} & \textbf{0.727} & \textbf{0.884} & \textbf{0.720} \\
\hline
\end{tabular}
\vspace{-10pt}
\end{table}

\subsection{Results}
In this section, we present the overall results and ablation studies. In addition, we also present the effectiveness of some of the modules such as DeepFake detection. We plan to evaluate other modules in detail in future work.

\paragraph{Overall Results:}
The model performs well on both Norwegian and English TikTok videos, with notable differences in class-wise behavior. For Norwegian content, the system achieves strong recall for \texttt{Checkworthy} instances (0.85) and high overall accuracy (0.88), indicating robust performance in identifying relevant claims. In contrast, the English dataset shows higher precision for \texttt{Checkworthy} (0.82) but lower recall (0.58), suggesting the model is more conservative in flagging English videos as checkworthy. Combined macro-averaged scores are comparable across languages (F1: 0.72 for Norwegian, 0.74 for English), highlighting the pipeline's cross-lingual generalizability with slightly better balance in the Norwegian case.

\subsection{Ablation Studies}
The ablation study shown in Table \ref{tab:ablation_deltas} demonstrates that textual modules are the most influential components in determining checkworthiness. The removal of the \textit{Transcript Verdict} and \textit{Buzzword} modules resulted in the largest decreases in recall and F1-score, highlighting the critical role of spoken content and ideological language. In contrast, excluding modules such as \textit{Weapon Detection}, \textit{Fact Check}, or \textit{Video-to-Text Verdict} had minimal impact, indicating their limited standalone contribution. Notably, the \textit{Weapon Detection} module slightly reduced overall performance, likely because such content appears infrequently; as a result, it was omitted from the final pipeline. These findings reinforce the system’s reliance on semantic and linguistic features rather than visual or metadata-based cues. Finally since removing individual modules does not show a huge drop in performance, the overall performance is attributed to contribution of modules.

\paragraph{DeepFake detection}
The models were evaluated in a zero-shot setting, without any fine-tuning on the target dataset. Among them, \textit{EfficientNET} outperformed all others across evaluation metrics, achieving an accuracy of 0.612 and a remarkably high precision of 0.992, indicating highly reliable positive predictions. However, its recall of 0.573 suggests it still misses nearly half of actual deepfakes. \textit{MesoNET} and \textit{Wvolf/ViT} perform very poorly.

\section{Demonstration}
\begin{figure}[ht!]
    \centering
    \includegraphics[width=0.44\textwidth]{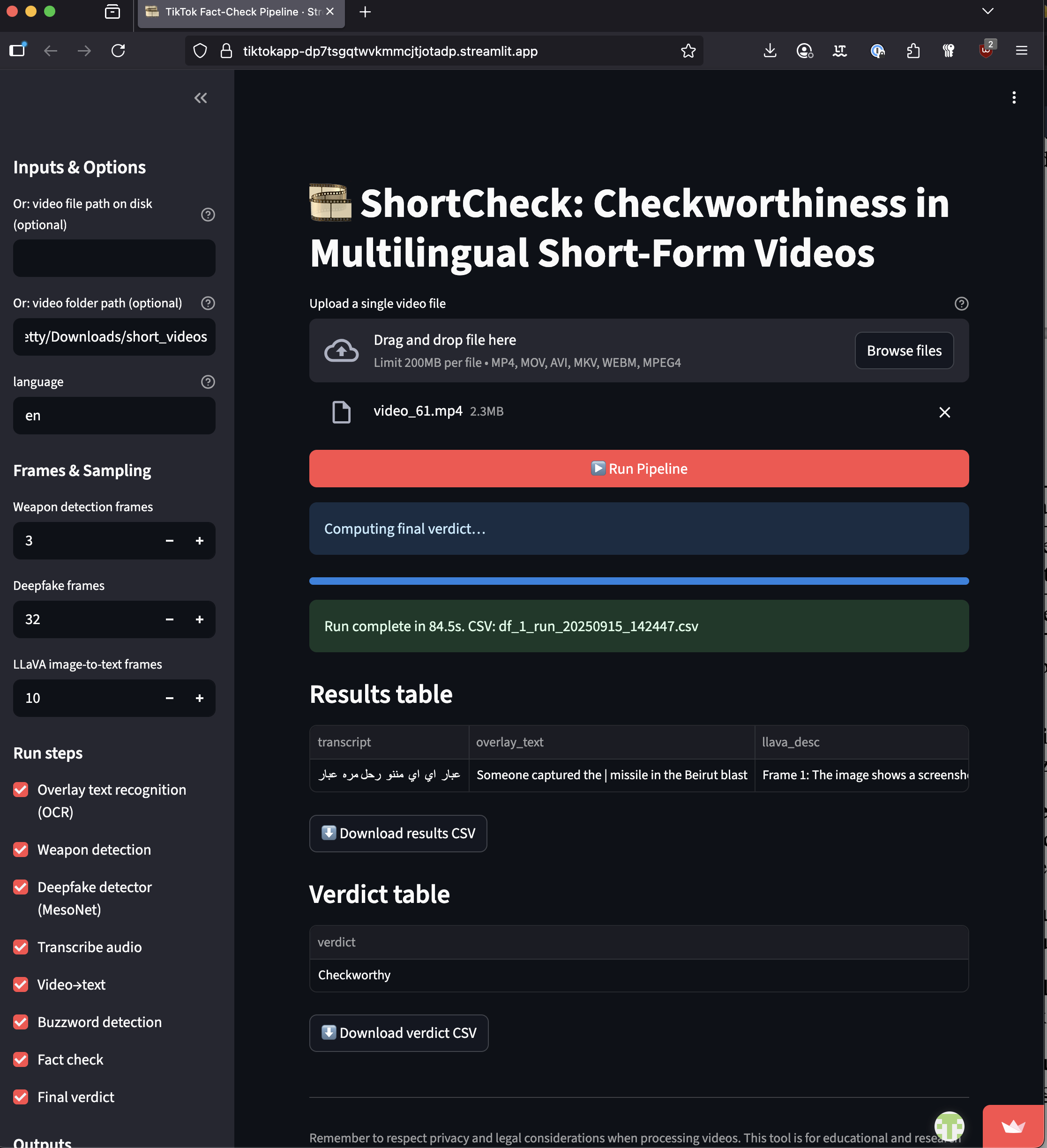}
    \caption{Demo interface for fact-checkers.}
    \label{fig:demo}
\end{figure}

Our demo system integrates the pipeline into a fact-checker-oriented web interface (Figure~\ref{fig:demo}).

\paragraph{Features.}
\begin{itemize}[noitemsep]
    \item Upload TikTok videos or paste URLs.
    \item Visualize intermediate module outputs (transcripts, OCR text, visual captions).
    \item Inspect final classification with explanation of contributing factors.
    \item Retrieve linked fact-checks via Factiverse API.
    \item Explore errors through confusion matrix visualizations.
\end{itemize}

\section{Conclusion and Future Work}
\label{sec:concl}
We presented 
\system, a modular, multilingual pipeline for detecting checkworthy content in short-form videos. The system integrates multiple modalities: text, audio, video, and image, and achieves strong performance on manually annotated TikTok datasets in both Norwegian and English. Through ablation studies, we showed that transcript and ideological language signals contribute most to checkworthiness, while visual features like deepfake detection and object recognition offer limited standalone utility.

In future work, we plan to extend \system~with full fact-checking while testing on more datasets, including multilingual video content from X and YouTube. We will evaluate individual modules in greater depth, improve robustness in noisy or low-resource settings, and integrate feedback from fact-checkers to refine outputs and interpretability. Finally, we aim to explore advanced multimodal fusion techniques beyond rule-based aggregation to boost accuracy and generalization.

\newpage
\bibliography{conf20yy-xxx}

\end{document}